\documentclass[letterpaper,10pt,conference]{ieeeconf}
\IEEEoverridecommandlockouts
\usepackage[T1]{fontenc}
\usepackage{times,graphicx,amsmath,amssymb,booktabs,array,cite,url,balance}

\usepackage{algorithm,algpseudocode}

\title{\LARGE\bf Kintsugi-VLA: Turning Failed Robot Rollouts into Recovery Data through Interventional Recoverability}

\author{%
    Ivan Snegirev*,
    Elizaveta Semenyakina*,
    \thanks{*These authors contributed equally to this work.}
    Dmitrii Maliukov, \\
    Miguel Altamirano Cabrera
    and Dzmitry Tsetserukou%
    \thanks{The authors are with the Intelligent Space Robotics Laboratory,
    Skolkovo Institute of Science and Technology. 
    {\tt\small \{ivan.snegirev, elizaveta.semenyakina, dmitrii.maliukov,
    m.altamirano, d.tsetserukou\}@skoltech.ru}}
}

\usepackage{xcolor}
\definecolor{Kold}{RGB}{135,45,45}
\definecolor{Knew}{RGB}{0,68,115}
\newif\ifKreview
\Kreviewfalse

\makeatletter
\newcommand{\Kignorelabels}{%
  \let\label\@gobble
  \let\label@in@display\@gobble
  \let\ltx@label\@gobble}
\makeatother

\begin{document}
\bstctlcite{IEEEexample:BSTcontrol}

\maketitle
\thispagestyle{plain}\pagestyle{plain}

\begin{abstract}
Simulation enables scalable training of Vision-Language-Action policies by using privileged experts to generate visual demonstrations without requiring every trajectory to be collected through manual teleoperation. However, such pipelines typically retain successful demonstrations while failed rollouts are discarded, even though they expose precisely the off-nominal states from which recovery must be learned.

We introduce Kintsugi-VLA, a framework for converting failed rollouts into targeted synthetic recovery data by exploiting exact state restoration and branching in simulation. For a fixed privileged expert, we define interventional recoverability as the probability of completing the original task after the simulator is restored to a given state, estimate it using adaptive Monte Carlo continuations with pointwise Wilson intervals, and characterize its non-monotonic evolution along failed trajectories. These estimates identify an observed terminal low-recoverability frontier---the point after which measured recoverability remains below a threshold---which is then used to select informative recovery starting states.

In a simulated Franka manipulation task, targeted recovery data yield aggregate SmolVLA recovery success of 34.6\% and 38.4\% under difficulty- and frame-budget matching, respectively, 5.8 and 6.7 percentage points above uniform sampling within the same recovery window. The same ordering is observed under disturbed end-to-end execution and shifted clutter and physics conditions, while clean-task success decreases from 76.8\% to 74.7\%. Kintsugi-VLA demonstrates how failed simulator rollouts can be transformed from discarded experience into structured recovery-training data through direct interventional measurement.

\end{abstract}

\section{INTRODUCTION}
Vision-Language-Action (VLA) models map multimodal observations and language instructions to robot actions and have become a promising approach to general-purpose manipulation~\cite{kim2024openvla,shukor2025smolvla}. Simulation and automated data-generation pipelines can reduce the amount of manual robot data collection by using privileged controllers to generate visual demonstrations at scale~\cite{mandlekar2023mimicgen,nasiriany2024robocasa,lum2025dextrah}. In our setting, a privileged simulation expert generates both nominal and recovery demonstrations through the same virtual camera interface used to train the VLA, without human teleoperation.

Success-only demonstration collection, however, leaves failed rollouts unused for imitation even though they contain states that are poorly represented by nominal trajectories. Missed grasps, unintended object motion, collisions, or stalled progress can move the policy away from its training distribution. Rather than imitating the failed actions themselves, we use these rollouts as sources of states from which new successful recovery demonstrations can be generated.

Existing work addresses failure through detection, recovery-data collection, or inference-time correction~\cite{gu2025safe,dai2024racer,shin2026b2ff,zhang2026core,zhao2026flare,liu2026liberorecover}. We focus on an upstream question: \emph{given a failed simulator rollout, from which states should recovery demonstrations be collected?} The final failed state may already be difficult to recover from, whereas states selected too early may largely repeat nominal behavior. Simulation provides a direct way to examine this trade-off: a saved state can be restored and the same privileged expert can be executed from it repeatedly.

\begin{figure}[t]
  \centering
  \includegraphics[width=\columnwidth]{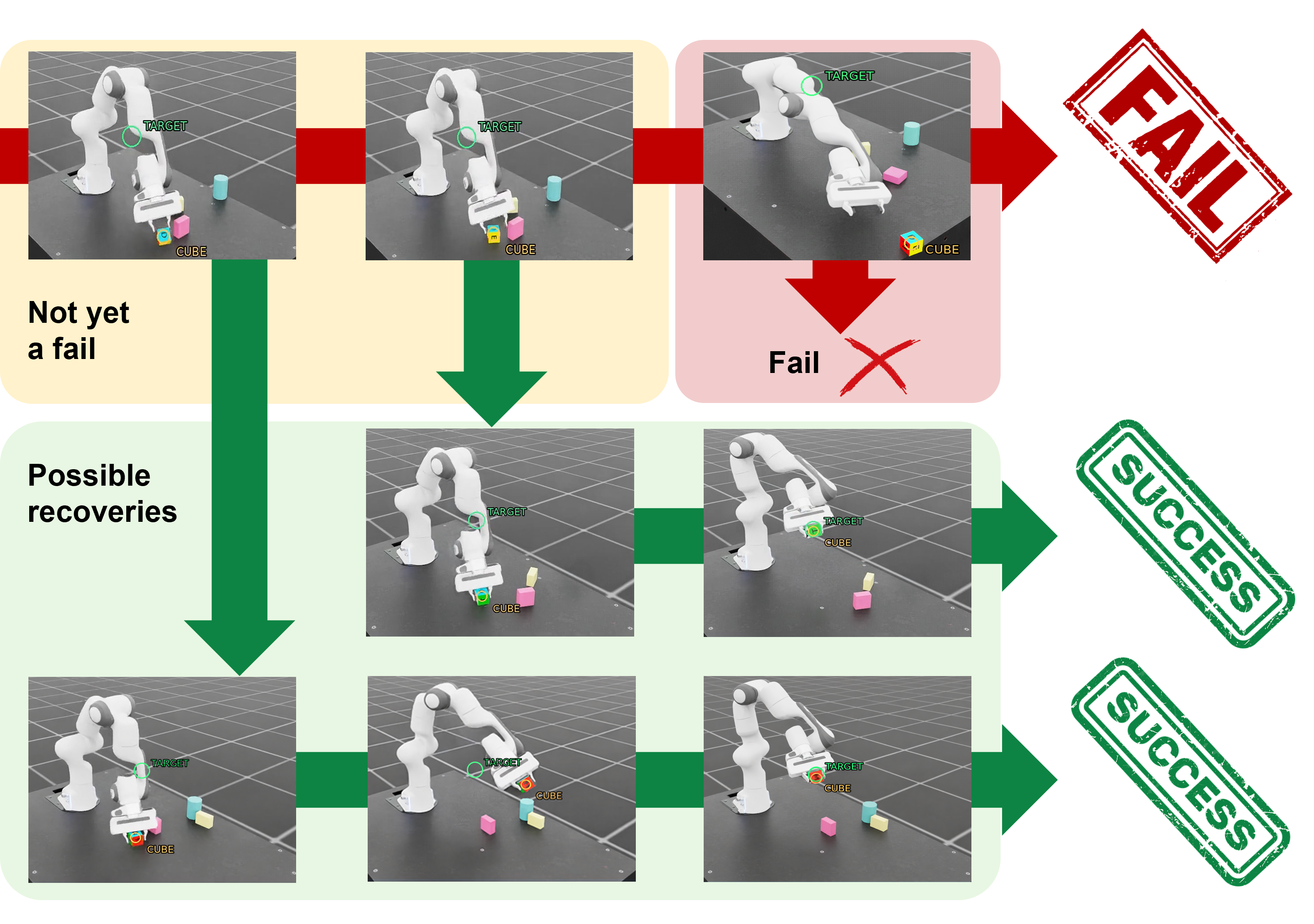}
  \caption{Failed rollouts as recovery-data sources. Kintsugi-VLA restores saved simulator states and repeatedly tests expert continuations to identify states suitable for synthetic recovery demonstrations.}
  \label{fig:main}
\end{figure}

Kintsugi-VLA uses this capability to measure expert-relative recoverability along failed trajectories. For each candidate checkpoint, we restore the simulator state and repeatedly transfer control to a fixed recovery expert. The fraction of successful continuations estimates
\(
p_{\mathrm{rec}}^{E}
\),
the probability that the specified expert completes the original task under the specified restoration and continuation protocol. These measurements characterize recovery structure and define candidate regions from which complete successful expert continuations are recorded as synthetic recovery demonstrations for VLA adaptation.

Our contributions are:
\begin{itemize}
    \item a restore-and-branch procedure that turns failed simulator rollouts into recovery-training data;
    \item an analysis of expert-relative recoverability, including uncertainty, non-monotonicity, recovery islands, and expert dependence;
    \item closed-loop SmolVLA evaluation under matched recovery-difficulty and visual-data budgets, where targeted selection achieves 34.6\% and 38.4\% recovery success versus 28.8\% and 31.7\% for uniform selection.
\end{itemize}

\section{RELATED WORK}

\textbf{Synthetic and corrective data generation.}
Simulation and automated data-generation methods can scale robot-learning datasets while reducing manual demonstration effort~\cite{mandlekar2023mimicgen,nasiriany2024robocasa,lum2025dextrah}. IntervenGen generates corrective intervention data to improve imitation-policy robustness~\cite{hoque2024intervengen}, while RaC fine-tunes policies using recovery-and-correction trajectories collected around failed execution~\cite{hu2025rac}. Kintsugi-VLA addresses the same broader problem of exploiting non-nominal experience, but uses simulator restoration to repeatedly test failed-rollout states under a privileged expert and uses the measured recoverability to select states for synthetic recovery-data collection.

\textbf{Failure detection and recovery.}
SAFE predicts VLA failure likelihood from internal representations~\cite{gu2025safe}, while RACER augments imitation learning with language-guided recovery trajectories~\cite{dai2024racer}. B2FF selects familiar visual milestones for inference-time recovery~\cite{shin2026b2ff}, CoRe uses counterfactual continuation and realignment~\cite{zhang2026core}, and FLARE combines failure monitoring with retry and reset behaviors~\cite{zhao2026flare}. LIBERO-RECOVER constructs recovery scenarios from failed executions and uses visual-language reasoning for failure localization and characterization~\cite{liu2026liberorecover}. In contrast, Kintsugi-VLA focuses on an upstream data-generation question: which states from a failed simulator rollout should be used to generate recovery demonstrations for subsequent VLA adaptation.

\section{METHOD}
\label{sec:method}

\begin{figure*}[t]
\centering
\includegraphics[width=\textwidth]{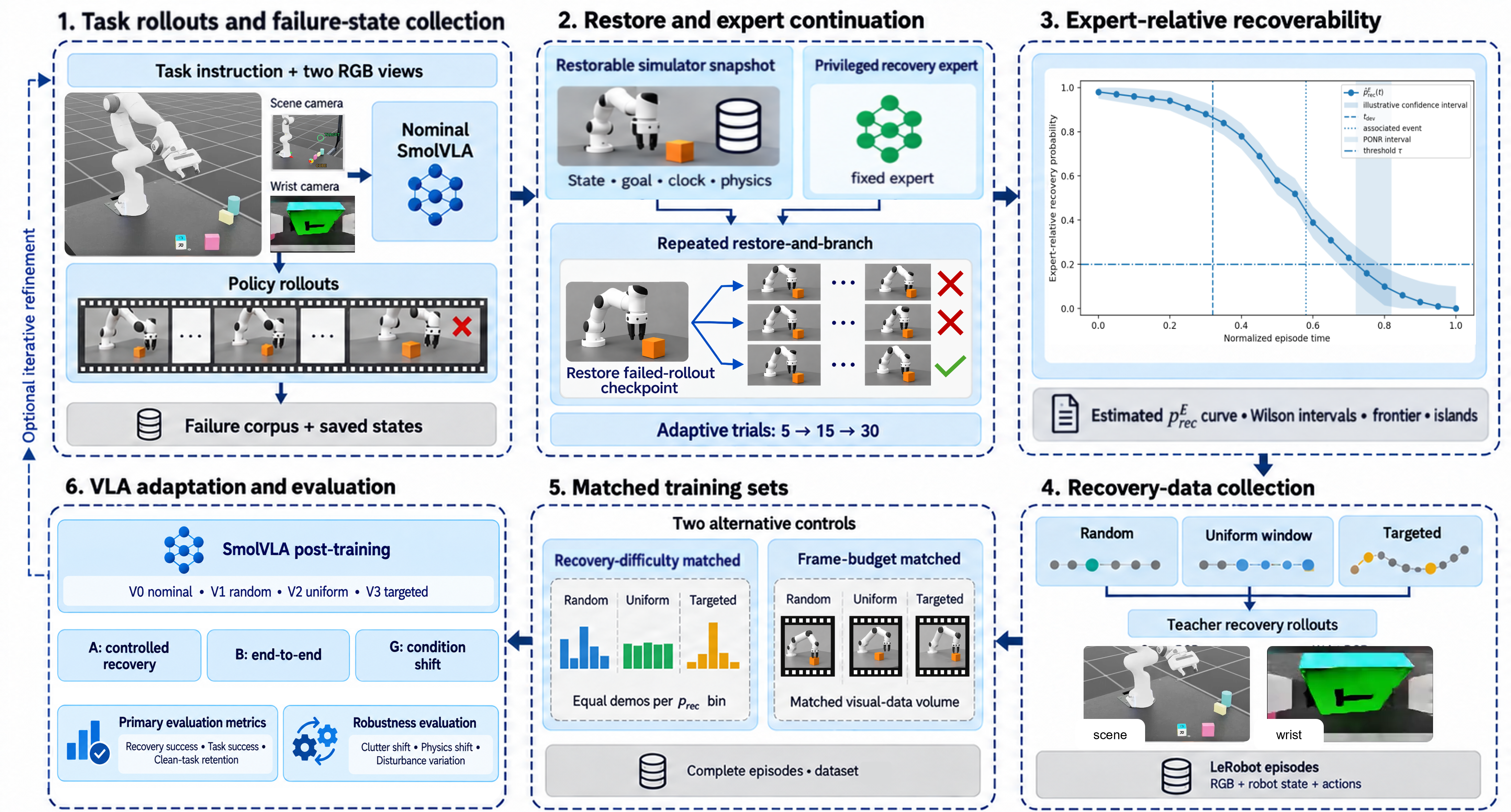}
\caption{Kintsugi-VLA method overview. Saved simulator states are restored and branched under a privileged expert. Recoverability measurements guide recovery-data collection, followed by difficulty- and frame-budget-matched VLA comparisons.}
\label{fig:method_overview}
\end{figure*}

Kintsugi-VLA converts failed simulator rollouts into recovery-training data by measuring which saved states remain recoverable for a privileged expert and collecting complete expert recovery demonstrations from selected states (Fig.~\ref{fig:method_overview}).

\subsection{Expert-relative recoverability}

Let $s_{e,i}$ be a simulator snapshot at checkpoint $t_i$ of the source episode $e$, $c$ the original task command, and $\pi_E$ a fixed recovery expert. The expert may use privileged simulator observations unavailable to the downstream VLA; its continuations serve as synthetic demonstrations, while the VLA receives only its permitted visual and robot-state inputs. Let's $H$ specify the continuation budget and $\mathcal{R}$ the restoration and termination protocol. We define:
\begin{equation}
p_{\mathrm{rec}}^E(s_{e,i};c,H,\mathcal{R},\nu)
=
\Pr_{\xi\sim\nu}
\left[
Y(s_{e,i},\pi_E,c,H,\mathcal{R},\xi)=1
\right],
\label{eq:recoverability}
\end{equation}
where $Y=1$ indicates completion of the original task before termination or budget exhaustion, and $\nu$ is the continuation-randomness distribution. Along a recorded trajectory, we abbreviate this quantity as $p_{\mathrm{rec}}^E(t_i)$.

This quantity is expert-relative rather than an intrinsic property of the physical state: it depends on the expert, task command, continuation horizon, restoration semantics, and continuation randomness. Each record retains the source episode, reset identity, simulator slot, checkpoint step, and branch seeds, linking repeated continuations to their trajectory for replay and dependence-aware analysis.

\subsection{Adaptive restore-and-branch estimation}

For checkpoint $i$, let $x_i$ of $n_i$ continuations succeed. We estimate $\hat p_i=x_i/n_i$ and retain a two-sided Wilson interval $[L_i,U_i]$~\cite{brown2001interval}:
\begin{equation}
[L_i,U_i]
=
\frac{
\hat p_i+z^2/(2n_i)
\ \pm\
z\sqrt{\hat p_i(1-\hat p_i)/n_i+z^2/(4n_i^2)}
}{
1+z^2/n_i
},
\label{eq:wilson_interval}
\end{equation}
with $z=1.95996$. Relative to threshold $\tau$, a checkpoint is below threshold when $U_i<\tau$, above threshold when $L_i\geq\tau$, and unresolved otherwise. The implemented budget ladder is $5\rightarrow15\rightarrow30$ total trials; refinement retains earlier outcomes and adds unused seeds. In the vectorized implementation, a saved batch is refined when any eligible failed-episode slot remains unresolved, while outcomes remain separate per slot.

Algorithm~\ref{alg:adaptive_measurement} specifies this procedure. The ladder allocates additional trials to uncertain estimates without assuming monotonicity. Wilson intervals are used as pointwise operational summaries and do not provide guaranteed 95\% coverage after adaptive stopping or simultaneous coverage across checkpoints; time-uniform inference requires a different construction~\cite{howard2021confidence}. With zero successes in 30 attempts, $U_i\simeq0.1135$, so this ladder cannot certify $U_i<0.1$.

\begin{algorithm}[t]
\caption{Adaptive expert-relative curve measurement}
\label{alg:adaptive_measurement}
\begin{algorithmic}[1]
\Require Snapshots with source IDs, expert $\pi_E$, threshold $\tau$, seed schedule
\State Initialize per-slot counts and branch records
\For{$b\in(5,15,30)$}
    \State Select all eligible batches if $b=5$; otherwise those with an unresolved eligible failed slot
    \State Add fresh-seed restore-and-branch trials up to total count $b$
    \State Update per-slot counts and Wilson intervals
\EndFor
\State For each failed episode, find its maximal terminal suffix satisfying $U_i<\tau$
\State Record frontier brackets, crossings, rises, and islands
\State \Return counts, seeds, intervals, and curve records
\end{algorithmic}
\end{algorithm}

\subsection{Stable frontier and non-monotonic structure}

For checkpoints $0,\ldots,m$, the observed terminal low-recoverability suffix begins at:
\begin{equation}
j^*=
\min\left\{
j:
U_i<\tau
\ \text{for every } i=j,\ldots,m
\right\}.
\label{eq:stable_suffix}
\end{equation}
When $j^*>0$, the frontier is localized to $(t_{j^*-1},t_{j^*}]$. If every checkpoint is below threshold, the earlier boundary is unknown; if the terminal checkpoint is unresolved or above threshold, no stable frontier is returned. Stability refers to the observed suffix, and the temporal bracket reflects checkpoint spacing.

\begin{figure}[t]
\centering
\includegraphics[width=\columnwidth]{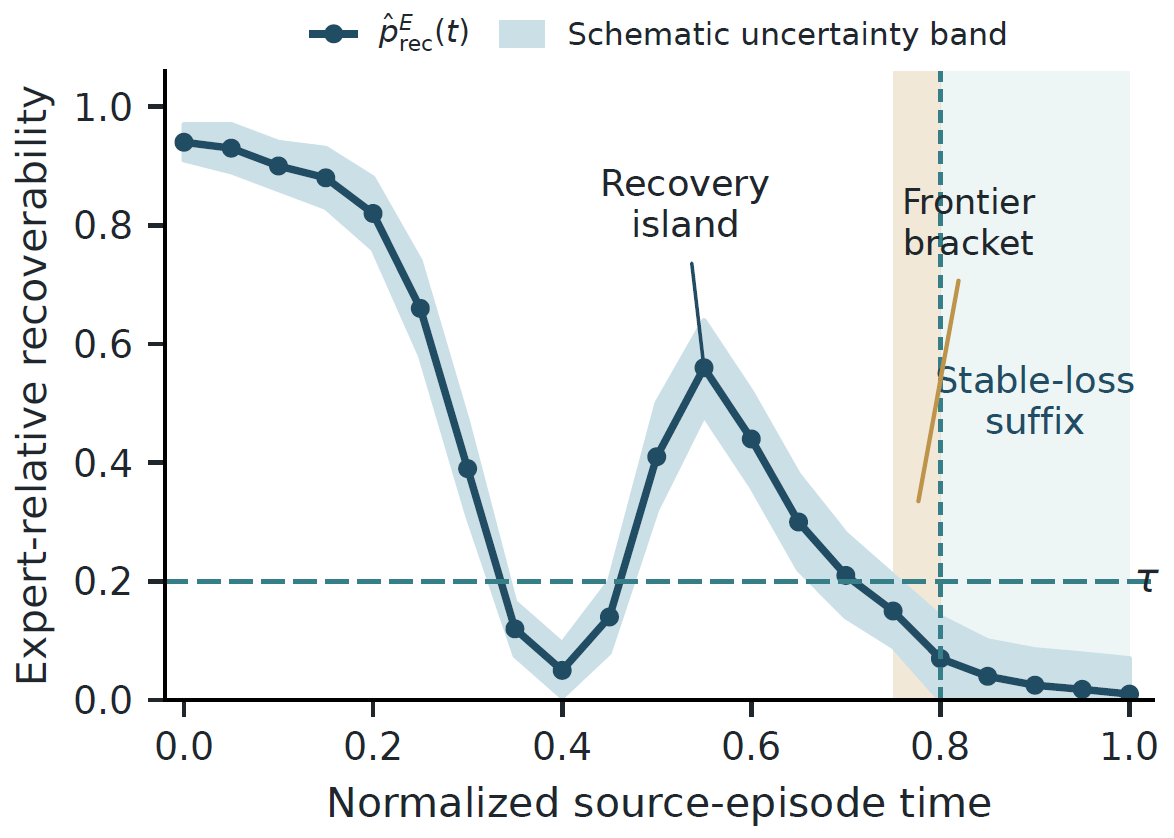}
\caption{Schematic expert-relative recoverability curve. A recovery island follows an earlier threshold crossing, while the frontier precedes the terminal suffix whose upper interval endpoints remain below $\tau$. Shading is illustrative, not a simultaneous confidence band.}
\label{fig:recoverability_curve}
\end{figure}

We retain three complementary curve summaries. Threshold crossings are changes in $\mathbb{1}[\hat p_i\geq\tau]$. An operational non-monotonicity flag records an adjacent rise with $L_{i+1}>U_i$. Recovery islands are contiguous above-threshold point-estimate regions following an earlier drop below threshold. Thus, an island and a separated rise use different criteria. Fig.~\ref{fig:recoverability_curve} illustrates why the first crossing alone can be insufficient.

\subsection{Recovery-state selection}

After recoverability measurement, failed rollouts serve as sources for synthetic recovery demonstrations. The generator compares targeted selection, uniform-window selection, and random valid-state perturbations under a shared privileged-expert continuation and export procedure. Targeted and uniform selection use the same failed source trajectories but differ in how starting states are positioned within the candidate recovery window.

The deviation time $t_{\mathrm{dev}}$ is the earlier of departure from a nominal progress corridor and a progress plateau; it defaults to zero if neither detector fires. When a valid frontier $t_f>t_{\mathrm{dev}}$ exists, normalized recovery depth is:
\begin{equation}
d(t)=
\frac{t-t_{\mathrm{dev}}}
{t_f-t_{\mathrm{dev}}}.
\label{eq:recovery_depth}
\end{equation}
The targeted sampler uses depths $0.15$, $0.50$, and $0.85$. Uniform selection samples within the same window, while random selection applies task-defined perturbations to valid states rather than sampling recorded trajectory times.

The current selector uses the measured recoverability structure through the terminal frontier $t_f$; local recovery-probability values and detected recovery islands are retained for analysis but do not directly alter the selected depths. When no stable frontier is available, the generator falls back to an alternative boundary or the episode end, retaining the corresponding source tag.

\subsection{Matching, export and data quality}

Matching is applied to demonstration manifests because successful recovery trajectories can differ in length. \emph{Difficulty matching} equalizes demonstration counts across shared expert-recoverability bins, while \emph{frame-budget matching} selects complete episodes toward a pooled difficulty histogram under a common frame allowance. Unique starting states, synchronized frames, and simulator steps are accounted for separately from demonstration counts.

Each successful privileged-expert continuation becomes a synthetic recovery demonstration in the same visual observation space used for downstream VLA training. Exported observations contain scene and wrist RGB, permitted robot-state fields, and the task instruction; recoverability estimates, privileged expert observations, failure labels, and branch metadata remain offline. Dataset QA checks view distinction, synchronization, timestamps, termination, episode boundaries, manifest consistency, and loader compatibility. Acquisition accounting includes unsuccessful continuations as well as successful exports.

\section{FRANKA EXPERIMENTAL SETUP}
\label{sec:franka_setup}

\subsection{Task, controller, and observations}

Experiments use a simulated Franka Panda with a parallel gripper in a cube lift-and-transport task. Task success requires the cube-center height to exceed $0.06$\,m and the cube-to-goal distance to be below $0.05$\,m. This instantaneous criterion requires neither release nor a post-success hold. The control period is $0.02$\,s.

The privileged teacher is trained with PPO in Isaac Lab using the \texttt{rsl\_rl} implementation. The primary teacher is the seed-42 checkpoint used in the canonical nominal evaluation and Protocol G teacher study. It acts as the synthetic data generator: teacher actions are executed in simulation while synchronized scene and wrist RGB observations are recorded for downstream VLA training. The teacher may access a privileged 43D simulator observation unavailable to the visual policy. No human teleoperation is used for nominal or recovery-data collection.

The canonical nominal evaluation comprises seven reset batches of 32 environments, yielding 224 episodes with a 250-step timeout. Only the first complete episode in each slot is counted. All outcomes are retained, while only successful episodes enter the nominal training dataset; unsuccessful rollouts remain available as recovery-state sources. Five additional demonstrations used for visualization and post-success video are excluded from this denominator.

The downstream policy uses SmolVLA~\cite{shukor2025smolvla}. Its input consists of two normalized $224\times224$ RGB views, the task instruction, and an 8D state containing environment-local end-effector position, quaternion, and previous gripper action. The action interface is 7D. Privileged teacher observations are never exposed to SmolVLA, and action-chunk queues are reset between attempts. Protocol G recordings use synchronized scene and wrist images at 50\,Hz.

Teacher scores use the original camera profile with corrected goal-marker synchronization. The revised profile removes the end-effector visualization marker while preserving the goal marker, camera poses, and frozen states; observation, physics, and inference checks are performed separately.

\subsection{Teacher acceptance and historical attribution}

The failed-rollout source corpus contains 200 episodes from both the privileged PPO teacher and the base VLA, with 33 teacher + 167 VLA ratio. Failed trajectories are not imitated directly; they provide candidate states from which the privileged teacher generates successful recovery continuations.

The historical teacher audit separately tests expert dependence: independently trained teachers with seeds 42, 1234, and 2026 each attempt the same 256 frozen states five times. A 127-episode historical trajectory subset containing failures from both teacher and VLA rollouts is analyzed with Algorithm~\ref{alg:adaptive_measurement} at $\tau=0.2$, with sensitivity checks at $0.1$ and $0.5$. Both historical studies predate the revised Protocol G restoration procedure and are reported separately.

\subsection{Policy comparisons and evaluation protocols}

V0 uses nominal demonstrations only; V1, V2, and V3 additionally use random-perturbation, uniform-window, and targeted recovery data, respectively. All recovery variants share the nominal dataset, privileged-teacher continuation contract, and visual interface. Difficulty matching and frame-budget matching define separate Protocol A comparisons; training manifests record selected episodes, frame counts, optimization exposure, and training seeds.

\textbf{Protocol A: controlled recovery.}
Policies begin from 300 shared frozen in-distribution recovery states, with five attempts per state and at least three training seeds per variant. After initialization, the VLA controls the continuation without expert assistance under a common budget. Success requires completion of the original task. State IDs and evaluation seeds are paired across policies.

\textbf{Protocol B: end-to-end execution.}
Evaluation begins from nominal task initialization. A disturbed track measures task completion under a fixed disturbance schedule, while a clean track measures nominal performance after recovery-data adaptation. Recovery gains and clean-performance changes are reported separately.

\textbf{Protocol G: condition shift.}
Evaluation uses frozen clutter and physics shifts defined below. Privileged-teacher recovery and closed-loop V0--V3 evaluation use separate policy and observation records. Protocol G denotes shifted Franka conditions; G1 evaluation is reserved for future work.

\subsection{Protocol G candidate pool and frozen set}

Protocol G crosses clutter seeds 20261706--20261709 with two physics profiles (Table~\ref{tab:physics_profiles}), yielding eight conditions. Each scene contains four distractors and a target from the same object family at scale 1.23.

\begin{table}[t]
\centering\small
\setlength{\tabcolsep}{3pt}
\caption{Protocol G physics profiles. Friction coefficients and mass multipliers are dimensionless; target scale is 1.23.}
\label{tab:physics_profiles}
\begin{tabular}{lccc}
\toprule
Profile & Static friction & Dynamic friction & Mass multiplier\\
\midrule
High friction & 1.25--1.35 & 1.05--1.15 & 1.00--1.20\\
Heavy object & 0.70--0.90 & 0.55--0.75 & 1.55--1.70\\
\bottomrule
\end{tabular}
\end{table}

Candidate collection uses 16 environments and 160 control steps per condition, saving snapshots every 20 steps. It includes intermediate teacher states and two cube XY/yaw perturbations in addition to each original state. The resulting pool contains 3,016 states, each evaluated with 15 calibration continuations, for 45,240 attempts.

\begin{figure}[t]
  \centering
  \includegraphics[width=\columnwidth]{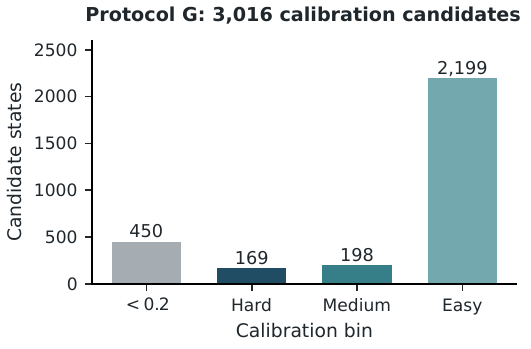}
  \caption{Protocol G calibration pool. Bin assignment uses 15 trials per state; the frozen evaluation set contains 30 states from each hard, medium, and easy bin.}
  \label{fig:g_candidates}
\end{figure}

Difficulty bins use the calibration estimate: hard $[0.2,0.5)$, medium $[0.5,0.8)$, and easy $[0.8,1]$. Within each condition and bin, a state-ID hash defines deterministic ordering. Round-robin selection across conditions yields 30 unique states per bin, or 90 frozen evaluation states in total. The 450 candidates below $0.2$ remain in the pool records but are excluded from this balanced set (Fig.~\ref{fig:g_candidates}).

The selected states are frozen before evaluation and receive five new trials under a schedule separate from calibration. States are never reselected using evaluation outcomes. Six conditions contribute 12 states each and two contribute nine, with equal bin counts within each condition. The resulting set balances recovery difficulty rather than reproducing the predominantly easy candidate distribution.

\subsection{Snapshot restoration and outcome accounting}

The revised evaluator restores robot and object state together with the target command, command timer, episode clock, previous actions, masses, inertias, and materials. Restored floating-point fields are checked to tolerance $2\times10^{-5}$. The original simulator slot is preserved because diagnostic repacking altered material values.

Five warm-up steps precede policy actions, with compensation for episode-clock and command-timer advancement. Recovery permits at most 240 policy steps while respecting the snapshot's remaining native episode budget. Only the first terminal event per slot is counted; padding slots and subsequent autoreset episodes are excluded. Warm-up success is recorded separately and assigned zero policy-action recovery time.

Recovery time excludes warm-up and is summarized over successful attempts. Failures retain elapsed time and termination codes. Because success terminates the episode, post-success failure is not observed.

\subsection{Evaluation units, uncertainty and leakage control}

Teacher results retain both successful-attempt and total-attempt counts; VLA comparisons report aggregate success rates and absolute percentage-point differences. We distinguish training seeds, source episodes, unique states, repeated continuations, and vectorized launches. Uncertainty in paired VLA comparisons depends on both independently trained policies and the shared evaluation-state and source-episode structure.

Recorded training/evaluation identifiers show no overlap. The earlier Protocol G audit compared all 3,016 candidate records with 602 training-manifest records and found no shared episode IDs or seeds. Those historical training records did not contain state IDs, object IDs, scene parameters, or source-snapshot hashes, so the audit cannot exclude overlap beyond the identifiers that were recorded.

\section{EXPERIMENTAL EVALUATION AND RESULTS}
\label{sec:results}

\subsection{Nominal teacher competence}
The canonical teacher succeeds in 223/224 nominal episodes (99.55\%); the timeout remains in the denominator. All episodes retain traces and both camera streams, while only successful episodes enter the nominal dataset. This high nominal competence supports its use as a synthetic demonstration generator but does not imply comparable recovery from off-nominal states.

\subsection{Historical trajectory structure}
\begin{figure}[t]
  \centering
  \includegraphics[width=\columnwidth]{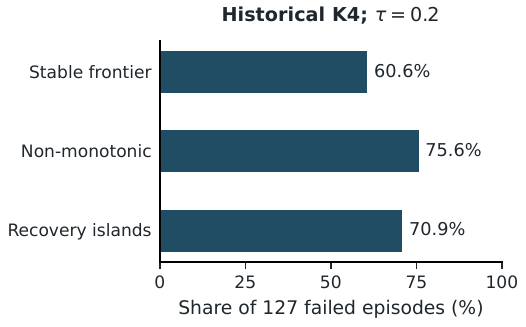}
  \caption{Historical trajectory structure across 127 failed episodes at $\tau=0.2$. Categories overlap; non-monotonicity uses the separated-rise criterion in Sec.~\ref{sec:method}.}
  \label{fig:k4_structure}
\end{figure}

We next ask whether failed simulator rollouts contain multiple recovery opportunities rather than a single failure point. The historical study analyzes 127 failed source episodes from the privileged teacher and base VLA. At $\tau=0.2$, 77 episodes (60.6\%) have an observed stable frontier, 75.6\% show non-monotonicity, and 90 episodes (70.9\%) contain recovery islands (Fig.~\ref{fig:k4_structure}). These properties overlap, so a first threshold crossing alone can miss later recoverable regions.

Conditional on a detected frontier, its normalized position has median 0.856 and 5th/95th percentiles 0.752/0.960, with a 10-control-step localization bracket. Curves average 3.37 threshold crossings. Measurement requires 2,825 vectorized branch launches and 17,779\,s of branching time; 93 of 100 saved batched snapshots reach the 30-trial stage.

Frontier counts are 0, 77, and 126 at thresholds $0.1$, $0.2$, and $0.5$. The zero at $0.1$ is constrained by the Wilson resolution limit. These historical summaries have not been recomputed with the revised Protocol G restoration procedure.

\subsection{Independent-teacher audit}

To test whether expert-relative recoverability is consistent across nominally competent teachers, three historical teachers are evaluated on the same 256 saved states. All achieve 100\% nominal acceptance success, while probe recovery ranges from 49.2\% to 54.8\% (Table~\ref{tab:teacher_acceptance}); each state receives five attempts per teacher.

\begin{table}[t]
\centering\small
\caption{Historical teacher acceptance and recovery probe. Each probe uses five attempts from 256 shared states.}
\label{tab:teacher_acceptance}
\begin{tabular}{lcc}
\toprule
Training seed & Nominal success & Probe recovery\\
\midrule
42   & 200/200 & 630/1280 (49.2\%)\\
1234 & 201/201 & 701/1280 (54.8\%)\\
2026 & 201/201 & 689/1280 (53.8\%)\\
\bottomrule
\end{tabular}
\end{table}

\begin{figure}[t]
  \centering
  \includegraphics[width=\columnwidth]{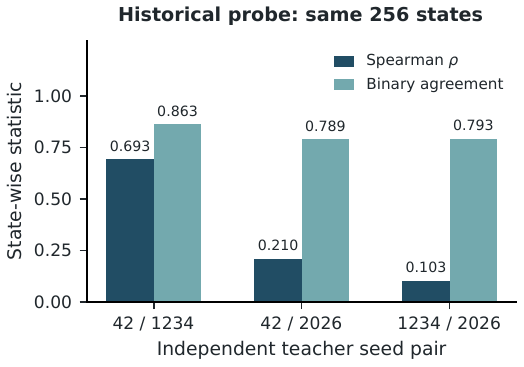}
  \caption{Teacher dependence on the same 256-state probe. Bars show state-wise Spearman correlation and binary agreement at $\tau=0.2$.}
  \label{fig:teacher_agreement}
\end{figure}

State-wise recovery rankings differ substantially across teachers. For pairs 42/1234, 42/2026, and 1234/2026, Spearman correlations are 0.693, 0.210, and 0.103, while binary agreement at $\tau=0.2$ is 0.863, 0.789, and 0.793 (Fig.~\ref{fig:teacher_agreement}). Mean absolute differences in estimated recovery probability are 0.205, 0.366, and 0.388, showing that coarse threshold agreement can mask large differences in state ordering.

The disagreement confirms that $p_{\mathrm{rec}}^E$ is a teacher-relative signal rather than ground-truth physical recoverability. With five attempts per state, the observed differences also include measurement noise; without complete trajectories for every teacher, cross-expert frontier stability remains untested.

\subsection{Frozen Protocol G teacher evaluation}

Across eight shifted clutter--physics conditions, the teacher succeeds in 31/32 nominal-start episodes and 310/450 recovery attempts (68.89\%) from 90 frozen states (Table~\ref{tab:g_conditions}).

\begin{table}[t]
\centering\small
\caption{Protocol G teacher outcomes. HF: high friction; HO: heavy object. Recovery uses five attempts per frozen state.}
\label{tab:g_conditions}
\begin{tabular}{ccrcc}
\toprule
Clutter seed & Profile & States & Ordinary & Recovery\\
\midrule
20261706 & HF & 12 & 4/4 & 46/60\\
20261706 & HO & 12 & 4/4 & 40/60\\
20261707 & HF & 12 & 4/4 & 50/60\\
20261707 & HO & 12 & 3/4 & 43/60\\
20261708 & HF & 12 & 4/4 & 38/60\\
20261708 & HO & 12 & 4/4 & 40/60\\
20261709 & HF & 9  & 4/4 & 27/45\\
20261709 & HO & 9  & 4/4 & 26/45\\
\midrule
Total & & 90 & 31/32 & 310/450\\
\bottomrule
\end{tabular}
\end{table}

\begin{figure}[t]
  \centering
  \includegraphics[width=\columnwidth]{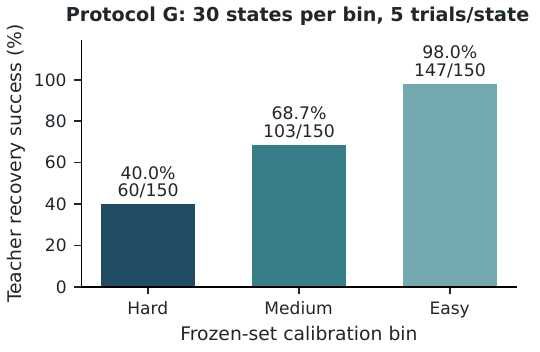}
  \caption{Protocol G teacher recovery on the frozen balanced set. Each difficulty bin contains 30 states with five evaluation attempts per state; bin assignment uses separate calibration trials.}
  \label{fig:g_recovery}
\end{figure}

Difficulty strata preserve their ordering under independent evaluation continuations: hard, medium, and easy states yield 60/150 (40.00\%), 103/150 (68.67\%), and 147/150 (98.00\%) successful recoveries, respectively (Fig.~\ref{fig:g_recovery}). This separation supports their use in constructing a balanced evaluation set.

The balanced set assigns equal weight to all three bins and excludes candidates below 0.2; its composition and restoration protocol therefore differ from the historical probe. Mean recovery time over the 310 successful attempts is 1.321\,s of policy actions. A state-bootstrap 95\% interval for aggregate success is 62.22--75.33\% using 10,000 resamples that keep all five attempts from each state together.

\subsection{Visual-data and inference checks}

Export checks confirm two-view synchronization, privileged-input isolation, and loader compatibility. Training-path and finite-action inference checks also pass for batch sizes one and 16 under the revised camera profile.

\subsection{Closed-loop VLA recovery and task performance}
\label{sec:vla_results}

When failed-rollout states are converted into privileged-expert recovery demonstrations, targeted selection yields the highest observed aggregate recovery success under both Protocol A matching schemes (Table~\ref{tab:vla_results}). V1 tests recovery-data collection from random valid-state perturbations, while V2 provides the stronger control: it uses the same frontier-defined window as V3 but samples recovery positions uniformly rather than at the fixed normalized depths of Sec.~\ref{sec:method}.

\begin{table}[t]
\centering\small
\setlength{\tabcolsep}{2pt}

\caption{Closed-loop VLA success (\%). Bold values indicate the highest aggregate rate in each column.}
\label{tab:vla_results}
\begin{tabular}{@{}lrrrrr@{}}
\toprule
 & \multicolumn{2}{c}{Protocol A} & \multicolumn{2}{c}{Protocol B} & Protocol G\\
\cmidrule(lr){2-3}\cmidrule(lr){4-5}\cmidrule(l){6-6}
Variant & Difficulty & Frames & Disturbed & Clean & Recovery\\
\midrule
V0 & 12.5 & 12.5 & 55.7 & \textbf{76.8} & 9.6\\
V1 & 25.9 & 28.8 & 58.6 & 75.8 & 22.1\\
V2 & 28.8 & 31.7 & 59.5 & 75.8 & 25.0\\
V3 & \textbf{34.6} & \textbf{38.4} & \textbf{63.4} & 74.7 & \textbf{28.8}\\
\bottomrule
\end{tabular}
\par\smallskip
\begin{minipage}{\columnwidth}\footnotesize
A: controlled recovery with difficulty- or frame-budget-matched data.
B: disturbed or clean end-to-end execution.
G: recovery under shifted clutter and physics.
V0: nominal-only; V1: random; V2: uniform-window; V3: targeted.
Between-training-seed dispersion is not reported for these aggregate rates.
\end{minipage}
\end{table}

Under difficulty matching, V3 reaches 34.6\% recovery success, compared with 25.9\% for V1 and 28.8\% for V2. Under frame-budget matching, the corresponding rates are 38.4\%, 28.8\%, and 31.7\%. V3 therefore yields observed aggregate gains of 5.8 and 6.7 percentage points over the uniform-window control.

The same ordering appears under disturbed end-to-end execution and shifted-condition recovery. Protocol B increases from 59.5\% for V2 to 63.4\% for V3, while Protocol G recovery increases from 25.0\% to 28.8\%. The latter remains well below the privileged teacher's 310/450 (68.89\%) recovery score, which is evaluated separately.

Recovery adaptation also introduces a nominal-performance trade-off. Clean Protocol B success is 74.7\% for V3, compared with 75.8\% for V2 and 76.8\% for V0. Thus, V3 gains 7.7 points over V0 under disturbed execution while losing 2.1 points on the clean task. These aggregate differences do not establish statistical significance or clean-task non-inferiority.

\subsection{Data efficiency}
\label{sec:data_efficiency}

A central motivation for exploiting failed simulator rollouts is to obtain useful recovery supervision without using every possible recovery continuation as training data. We therefore measure efficiency with respect to successful-demonstration and synchronized-frame budgets. Let $S_v(B)$ denote recovery success for variant $v$ at budget $B$. For recovery level $q$:
\begin{equation}
\begin{aligned}
B_v(q) &= \inf\{B:S_v(B)\geq q\},\\
R_q &= \frac{B_{\mathrm{uniform}}(q)}
{B_{\mathrm{targeted}}(q)} .
\end{aligned}
\label{eq:efficiency_ratio}
\end{equation}
Values $R_q>1$ indicate a smaller targeted-data budget on the corresponding resource axis.

\begin{table}[t]
\centering\small
\setlength{\tabcolsep}{4pt}
\caption{Interpolated normalized budgets required to reach recovery score $q$.}
\label{tab:data_efficiency}
\begin{tabular}{lrrrr}
\toprule
Budget axis & $q$ (\%) & Uniform & Targeted & $R_q$\\
\midrule
Demonstrations & 34 & 1.333 & 0.925 & 1.44\\
Frame records  & 37 & 1.333 & 0.884 & 1.51\\
\bottomrule
\end{tabular}
\end{table}

At 34\% recovery success, targeted selection uses 30.6\% fewer successful demonstrations than uniform selection; at 37\%, it uses 33.7\% fewer synchronized frame records. These results characterize training-data efficiency only and do not include the simulator compute required for restore-and-branch recoverability estimation.

\subsection{Diagnosis-conditioning extension}
\label{sec:diagnosis}

We additionally test whether explicit failure diagnosis improves targeted recovery training. Oracle diagnosis increases difficulty-matched recovery from 34.6\% to 37.4\%, while predicted diagnosis reaches 34.3\%, indicating that reliable semantic diagnosis can complement interventional state selection but is not required by the core Kintsugi-VLA pipeline.

\section{LIMITATIONS}
\label{sec:limitations}

\textbf{Dependence on the recovery expert.}
Privileged simulation does not make recoverability an objective state property. Recoverability labels depend on the chosen expert, continuation budget, and restoration protocol. The audit shows substantial ranking variation between nominally competent experts ($\rho=0.103$--$0.693$), partly confounded by five-trial estimation noise. Changing the expert may alter difficulty bins, selected states, and demonstrations. Cross-expert robustness of the VLA benefit and stable frontiers remains untested.

\textbf{Uncertainty of the VLA comparisons.}
Despite at least three training seeds per variant, Protocol A reports no per-seed dispersion. Statistical precision, significance, and clean-task non-inferiority therefore remain unresolved. Proper uncertainty estimation must account for both independently trained policies and the paired state/episode structure. The teacher bootstrap targets a different estimand and cannot quantify VLA uncertainty.

\textbf{Measurement and evaluation scope.}
The current experiments establish the data-generation concept only for one simulated Franka task with a fixed target family and scale under specified clutter and physics shifts. They do not establish that interventional recoverability provides the same selection advantage across tasks, embodiments, or different privileged-teacher training procedures. Hardware/G1, release-and-place behavior, and post-success stability remain untested. Finite checkpoints and trial budgets limit measurement resolution, and adaptive Wilson intervals lack time-uniform coverage guarantees. Historical cohorts also differ in restoration semantics, preventing direct longitudinal comparison.

Protocol G uses balanced difficulty bins and excludes candidates below $0.2$, so its aggregate recovery rate should not be interpreted as population-average performance. Bin separation establishes teacher-level difficulty ordering but not fine-grained calibration or direct transfer to VLA recovery difficulty. Leakage auditing is limited to the recorded identifiers available in the manifests.

\textbf{Selection, diagnosis, and total cost.}
The current selector uses fixed normalized depths and does not directly exploit local recovery-probability values or detected recovery islands. Island-aware selection and joint optimization of recovery points and measurement budget remain open. Reported data-efficiency results concern successful demonstrations and synchronized frames only; total savings remain unmeasured after including restoration, recoverability estimation, unsuccessful continuations, and training. Diagnosis conditioning measures information utility rather than physical-cause identification, and predicted diagnosis does not improve over neutral inputs.

We also do not experimentally compare interventional recoverability with a VLM-based failure-localization or recovery-difficulty estimator. The distinction from such approaches is therefore methodological rather than an empirical superiority claim.

\balance

\section{CONCLUSION}

We introduced Kintsugi-VLA, a framework for converting failed simulator rollouts into targeted synthetic recovery data. Rather than discarding unsuccessful executions or treating the final failure state as the natural recovery point, Kintsugi exploits simulator state access to restore candidate checkpoints and repeatedly evaluate them under a privileged expert. The resulting success frequency defines expert-relative recoverability and provides a direct interventional signal for organizing recovery-data collection.

The measured recovery structure is neither strictly monotonic nor independent of the recovery expert. Historical trajectories contain later recoverable regions after earlier drops in recoverability, while nominally competent teachers show substantial differences in state-wise recovery rankings. These findings support treating recoverability as an expert- and protocol-dependent quantity and motivate using the trajectory structure, rather than a single failure point, when selecting recovery starting states.

Closed-loop SmolVLA experiments show that targeted recovery-data selection yields higher observed aggregate recovery success than random and uniform-window controls. Relative to uniform-window selection, V3 gains 5.8 and 6.7 percentage points under difficulty- and frame-budget matching, respectively, and 3.9 and 3.8 points under disturbed end-to-end and shifted-condition evaluation. Targeted selection also reaches the reported recovery levels with 30.6\% fewer successful demonstrations and 33.7\% fewer synchronized frame records. These gains are accompanied by a 2.1-point reduction in clean-task success relative to nominal-only training, indicating a measurable trade-off between recovery adaptation and nominal performance.

The results establish a practical use for simulator intervention beyond generating successful demonstrations. Failed rollouts can themselves become structured training resources when the simulator is used to test where successful continuation remains possible and to regenerate successful expert behavior from those states. Within the evaluated task and expert, Kintsugi-VLA shows that interventional recoverability can serve as a useful signal for recovery-data curation while keeping privileged information outside the deployed VLA.

\section{FUTURE WORK}
Future work will (i) evaluate recovery on G1 under balance, contact, and whole-body constraints; (ii) extend recovery-data generation across tasks, object families, perturbations, observations, and recovery experts; (iii) jointly optimize recovery-state selection and measurement budgets, accounting for restoration, expert queries, unsuccessful continuations, and training cost; and (iv) improve automatic diagnosis-conditioned recovery for missed contact, slippage, collision, and joint-limit failures.
\bibliographystyle{IEEEtran}
\bibliography{references}
\end{document}